\newcommand{\spl}[1]{{\color{black} #1}}
\documentclass[sigconf]{acmart}

\AtBeginDocument{%
  }
\usepackage{multirow}
\usepackage{bbding}
\usepackage{tcolorbox}
\usepackage{mdframed}
\newmdenv[
  leftmargin=10em,              
  innerleftmargin=10pt,
  linecolor=gray!60!black,
  linewidth=2pt,
  topline=false,
  rightline=false,
  bottomline=false,
  innertopmargin=0pt,
]{linedpara}
\setcopyright{acmlicensed}
\copyrightyear{2025}
\acmYear{2025}
\acmDOI{XXXXXXX.XXXXXXX}

\acmISBN{978-1-4503-XXXX-X/18/06}

\begin{document}

\title{City-VLM: Towards Multidomain Perception Scene Understanding via Multimodal Incomplete Learning}


\author{Penglei Sun}
\orcid{0009-0005-2290-0944}
\affiliation{%
  \institution{The Hong Kong University of Science and Technology (Guangzhou)}
  \city{Guangzhou}
  \country{China}
  }
\email{psun012@connect.hkust-gz.edu.cn}
\authornote{Equal Contribution.}

\author{Yaoxian Song}
\orcid{0000-0002-8146-2236}
\affiliation{%
  \institution{Zhejiang University}
  \city{Hangzhou}
  \country{China}
  }
\email{songyaoxian@zju.edu.cn}
\authornotemark[1]

\author{Xiangru Zhu}
\orcid{0000-0001-7308-3642}
\affiliation{%
  \institution{Fudan University}
  \city{Shanghai}
  \country{China}
  }
\email{xrzhu19@fudan.edu.cn}

\author{Xiang Liu}
\orcid{0009-0003-2639-1804}
\affiliation{%
  \institution{The Hong Kong University of Science and Technology (Guangzhou)}
  \city{Guangzhou}
  \country{China}
  }
\email{xliu886@connect.hkust-gz.edu.cn}

\author{Qiang Wang}
\orcid{0000-0002-2986-967X}
\affiliation{%
  \institution{Harbin Institute of Technology (Shenzhen)}
  \city{Shenzhen}
  \country{China}
  }
\email{qiang.wang@hit.edu.cn}
\authornote{Corresponding author.}

\author{Yue Liu}
\affiliation{%
  \institution{Terminus Technologies Co., Ltd.}
  \city{Chongqing}
  \country{China}
  }
\email{liu.yue@tslsmart.com}
\authornotemark[2]

\author{Changqun Xia}
\affiliation{%
  \institution{Pengcheng Laboratory}
  \city{Shenzhen}
  \country{China}
  }
\email{xiachq@pcl.ac.cn}

\author{Tiefeng Li}
\orcid{0000-0003-0265-3454}
\affiliation{%
  \institution{Zhejiang University}
  \city{Hangzhou}
  \country{China}
  }
\email{litiefeng@zju.edu.cn}

\author{Yang Yang}
\orcid{0000-0003-0608-9408}
\affiliation{%
  \institution{The Hong Kong University of Science and Technology (Guangzhou)}
  \city{Guangzhou}
  \country{China}
  }
\email{yyiot@hkust-gz.edu.cn}

\author{Xiaowen Chu}
\orcid{0000-0001-9745-4372}
\affiliation{%
  \institution{The Hong Kong University of Science and Technology (Guangzhou)}
  \city{Guangzhou}
  \country{China}
}
\email{xwchu@ust.hk}
\authornotemark[2]

\newcommand{\syx}[1]{{\color{black} #1}}

\begin{abstract}

Scene understanding enables intelligent agents to interpret and comprehend their environment.
While existing large vision-language models (LVLMs) for scene understanding have primarily focused on indoor household tasks, they face two significant limitations when applied to outdoor large-scale scene understanding. 
First, outdoor scenarios typically encompass larger-scale environments observed through various sensors from multiple viewpoints (e.g., bird view and terrestrial view), while existing indoor LVLMs mainly analyze single visual modalities within building-scale contexts from humanoid viewpoints. 
Second, existing LVLMs suffer from missing multidomain perception outdoor data and struggle to effectively integrate 2D and 3D visual information.  
To address the aforementioned limitations, we build the first multidomain perception outdoor scene understanding dataset, named \textbf{\underline{SVM-City}}, deriving from multi\textbf{\underline{S}}cale scenarios with multi\textbf{\underline{V}}iew and  multi\textbf{\underline{M}}odal instruction tuning data. 
It contains $420$k images and $4, 811$M point clouds with $567$k question-answering pairs from vehicles, low-altitude drones, high-altitude aerial planes, and satellite.
To effectively fuse the multimodal data in the absence of one modality, we introduce incomplete multimodal learning to model outdoor scene understanding and design the LVLM named \textbf{\underline{City-VLM}}. 
Multimodal fusion is realized by constructing a joint probabilistic distribution space rather than implementing directly explicit fusion operations (e.g., concatenation). 
Experimental results on three typical outdoor scene understanding tasks show City-VLM achieves $18.14 \%$ performance surpassing existing LVLMs in question-answering tasks averagely. 
Our method demonstrates pragmatic and generalization performance across multiple outdoor scenes.
Our project is available on our  website\footnote{\url{https://sites.google.com/view/cityvlm/}}.

\end{abstract}

\ccsdesc[500]{Computing methodologies~Natural language processing}
\ccsdesc[500]{Computing methodologies~Scene understanding}

\keywords{multimodal question answering, scene understanding, 3D}


\maketitle

\section{Introduction}

Scene understanding involves enabling agents to recognize and interpret the semantic information of objects within their surrounding environment~\cite{chen2024towards}, which is a fundamental task for autonomous navigation~\cite{deruyttere2019talk2car,yu2023brain,song2024scene}, robot manipulation~\cite{song2022human,sarch2023open,sun2024multi}, digital city~\cite{xiang2020learning}, etc.
Technically, it usually involves space-sky-land multidomain perception data in multimodal (e.g., image, point cloud) gathered from multiview observation (e.g., humanoid view, terrestrial view, and bird view) to profile cities at multiple scales~\cite{sola2020multi}.
Currently, large vision language models (LVLMs) are used to model scene understanding problems popularly, which are fed with visual-texture information and generate text descriptions of a situated environment for an agent~\cite{fei2022towards}. 
The existing research mainly investigates indoor scene understanding while LVLMs in outdoor scene understanding have not been explored systematically.

\begin{figure}[t]
  \centering
  \includegraphics[width=0.85\linewidth]{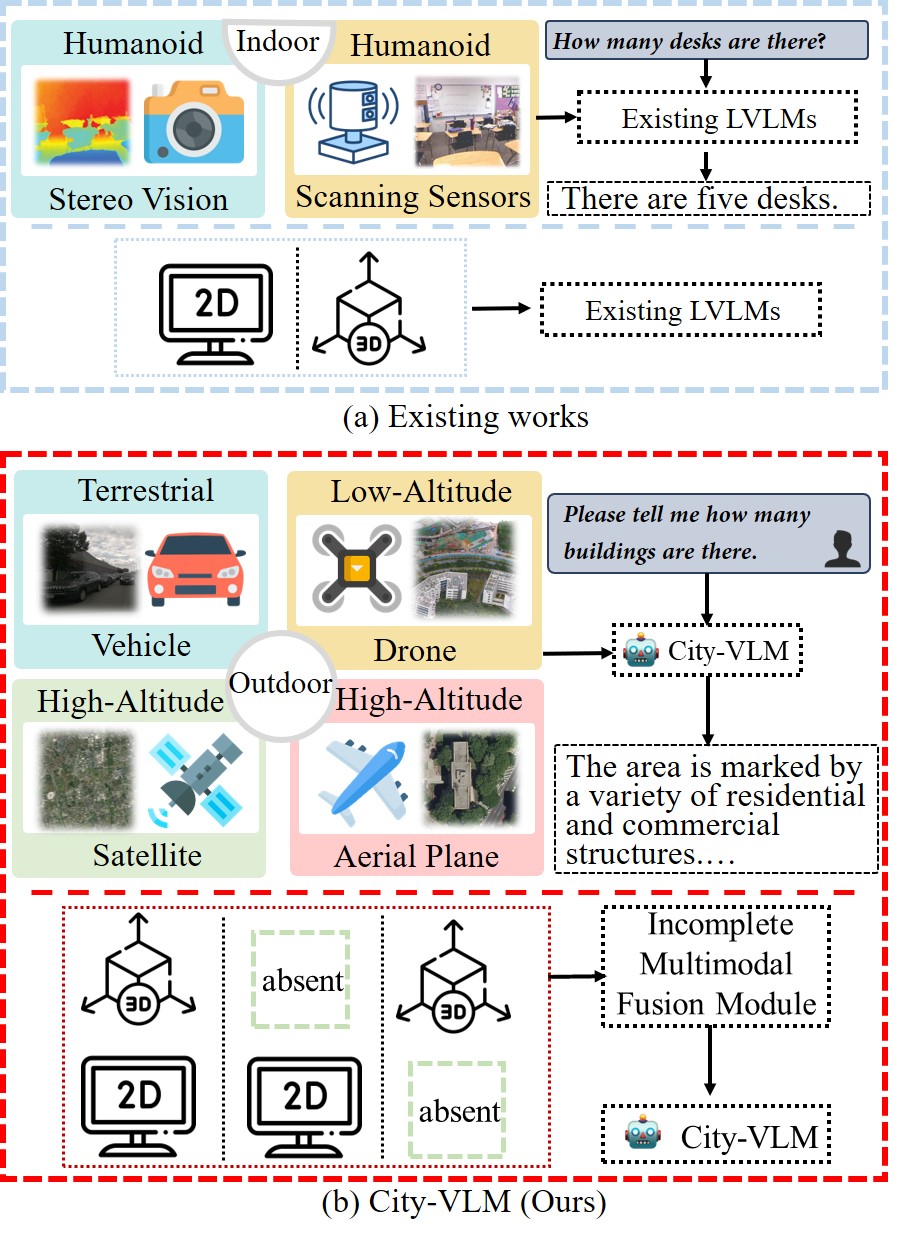}
  \caption{
Various large vision-language models (LVLMs) in scene understanding. (a) Existing works focus on \underline{indoor scenes} with single-scale visual data collected by limited stereo-vision cameras or scanning sensors from a humanoid view. They process a single visual modality (e.g., 2D or 3D data) at a time. 
(b) Our work City-VLM studies \textbf{multi-$\{$Scale, View, Modal$\}$} scene understanding \underline{outdoors}. 
City-VLM employs the Incomplete Multimodal Fusion Module (IMF-Module) to model the incomplete visual perception (e.g., the 2D data or the 3D data is absent).
  }
  \label{fig:overview}
\end{figure}

In indoor environments, as shown in Figure~\ref{fig:overview} (a), LVLMs integrate vision and language representations to respond to the context of indoor scenes~\cite{hong2024multiply,chen2024ll3da,fu2024scene,li20243dmit,huang2023chat}. 
The datasets in these studies, such as ScanNet~\cite{dai2017scannet, armeni20163d} and Matterport3D~\cite{chang2017matterport3d}, are primarily collected using portable devices equipped with scanning sensors or stereo-vision cameras. 
These LVLMs are often trained on downstream tasks like question-answering (QA) related to household activities at the building scale from the humanoid viewpoint, where they generally process a single visual modality (e.g., 2D or 3D data) at a time.
In contrast, outdoor scenes are usually constructed through space-sky-land multidomain perception data collected from sources including terrestrial vehicle cameras~\cite{caesar2020nuscenes,geiger2012we}, low-altitude drones~\cite{yang2023urbanbis,hu2022sensaturban}, and high-altitude aircraft or satellites~\cite{wang2loveda}.
However, existing LVLM research for the outdoors has not fully integrated multiscale, multiview, and multimodal visual data, nor does it effectively handle the simultaneous processing of these diverse data~\cite{yang2023lidar,cao2024maplm}.

\begin{figure*}[t]
  \centering
  \includegraphics[width=0.8\linewidth]{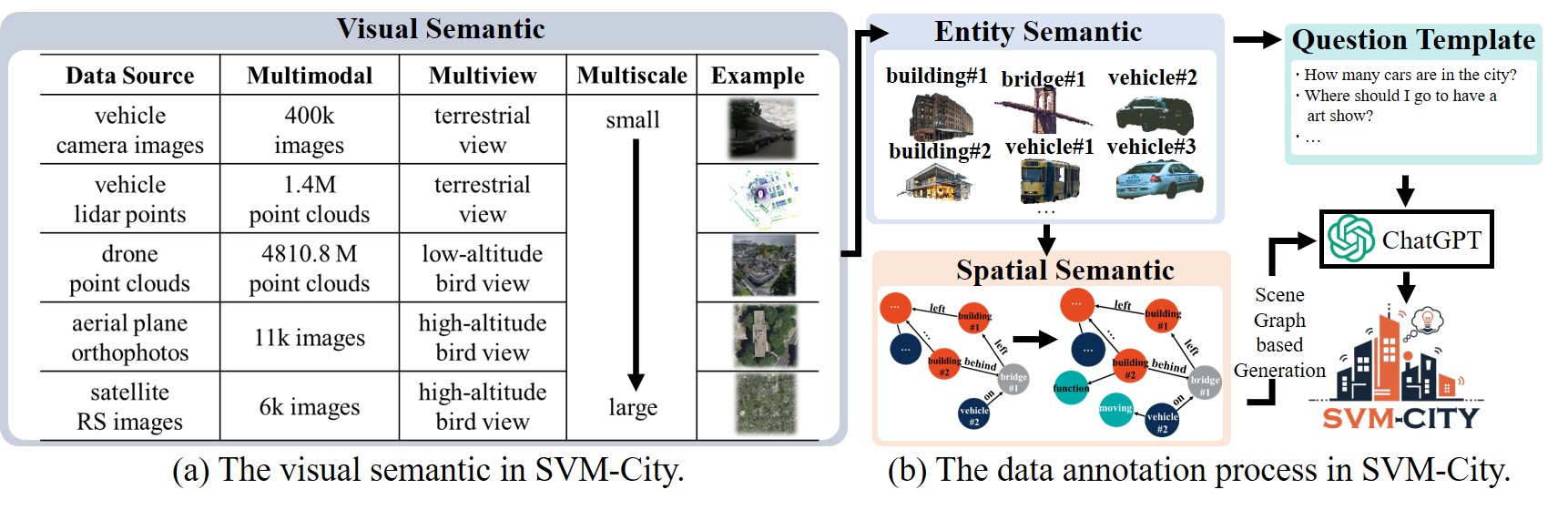}
  \caption{
  The overview of SVM-City: (a) the visual semantic from SVM-City and (b) the data annotation process applied to the SVM-City dataset.
  }
  \label{fig:pieline_annotation}
\end{figure*}

To address these challenges, we explore the LVLMs in outdoor scene understanding by constructing a novel \underline{dataset} SVM-City and the first outdoor \underline{LVLM model} City-VLM,  as shown in Figure~\ref{fig:overview} (b).
\textbf{For dataset design}, we propose the instruction tuning datasets from the multi\textbf{S}cale outdoor city-level scene based on multi\textbf{V}iew observation and multi\textbf{M}odal data, called \textbf{SVM-City}.
We collect multiscale visual data including high-altitude satellite remote sensing (RS) images, high-altitude aerial orthophotos, low-altitude drone point clouds, and terrestrial vehicle camera and lidar points. 
We design a question taxonomy that includes five question types to address common queries related to outdoor scene understanding based on the spatial question categories in cognitive science research~\cite{hayward1995spatial}.
Then we propose the automatic data annotation based on the ChatGPT and existing segmentation method.
SVM-City contains $420$k images and $4,811$M point clouds with $567$k QA pairs.

For model design, we propose an LVLM, named \textbf{City-VLM}, using multimodal data from multiview observation for the multiscale outdoor city-level scene (SVM-City) based on incomplete multimodal learning inspired by ~\citet{wei2024robust}.
In contrast to conventional multimodal fusion methods\cite{ngiam2011multimodal}, our City-VLM attempts to construct a joint probabilistic distribution space for the multimodal input through the VAE-based~\cite{kingma2013auto} \underline{\spl{Incomplete Multimodal Fusion Module}} (IMF Module), as shown in Figure~\ref{fig:overview} (b). 
Specifically, it is common that the partial input visual modalities are missing in practical outdoor environments. 
For example, in high-altitude situations, only 2D remote sensing images are available while 3D data is absent. 
To address it, a shared visual representation of incomplete visual perception is obtained by sampling from a probabilistic distribution based on the only available 2D image data. 
Experimental results indicate that our method outperforms the existing LVLMs $18.14 \%$ averagely in three outdoor question-answering (QA) tasks.

Our main contributions can be summarized as follows:

$\bullet$ We are the first to elaborately investigate outdoor scene understanding on multiscale, multiview, and multimodal city-level using LVLM, making it perform robustly in real scenarios.

$\bullet$ We propose \textbf{SVM-City}, the first outdoor city-level multi\textbf{\underline{S}}cale, multi\textbf{\underline{V}}iew, and multi\textbf{\underline{M}}odal instruction tuning dataset. It consists of $420$ thousand images and $4810.8$ million point clouds with $567$ thousand question-answering (QA) pairs.

$\bullet$ We design an incomplete-learning-based LVLM, named \textbf{City-VLM}. An \spl{Incomplete Multimodal Fusion Module} (IMF Module) is designed to construct a joint probabilistic distribution space over 2D and 3D modalities, which enhances the visual representation for LVLM in case of corrupted sensor modalities.


$\bullet$ 
Experiments are performed on three typical outdoor tasks including object recognition, spatial reasoning, functionality prediction, and logicality inference tests over \underline{road}-\underline{low altitude}-\underline{high altitude} viewpoints. 
Results show that our method has obvious advantages over existing LVLMs with a $18.14 \%$ averagely. 
Remarkably, our method outperforms low-altitude QA, advanced of existing LVLMs only skilled at humanoid viewpoints by $30 \%$ averagely.


\section{Related Work}

\subsection{Datasets in Scene Understanding}
RGB-D datasets are extensively utilized in indoor environments, primarily collected through portable scanning sensors integrated into handheld devices such as iPhones and iPads. 
Datasets like ScanNet~\cite{dai2017scannet,armeni20163d} and Habitat-Matterport~\cite{ramakrishnan2habitat} focus on indoor semantic segmentation, offering dense and detailed annotations of various 3D indoor objects. 
In contrast, outdoor scene understanding presents more challenges due to its complexity and large scale, leading researchers to collect multimodal outdoor datasets.
Datasets such as NuScenes~\cite{caesar2020nuscenes} and KITTI~\cite{geiger2012we} emphasize traffic scenes for autonomous driving, using lidar and vehicle-mounted cameras to capture multimodal data. 
Additionally, studies by ~\citet{yang2023urbanbis} and ~\citet{hu2022sensaturban} employ low-altitude drones to collect 3D point cloud data for urban scene reconstruction and segmentation. 
Furthermore, ~\citet{zhang2023vehicle} and ~\citet{su2021urban} explore satellite and aerial remote sensing imagery to inform urban planning decisions.

\subsection{Large Vision-Language Models}

With the advancement of Large Vision-Language Models (LVLMs)~\cite{alayrac2022flamingo,liu2024visual,li2023blip,dubey2024llama}, recent efforts have focused on adapting these models for visual understanding and reasoning tasks in scene comprehension. 
For indoor scene understanding, researchers\cite{hong2024multiply,chen2024ll3da,fu2024scene,li20243dmit,huang2023chat} propose models which integrate both language and 3D visual information from human input to enable understanding, reasoning, and planning in 3D indoor environments based on ScanNet~\cite{dai2017scannet} or Matterport3D~\cite{chang2017matterport3d}.
Researchers~\cite{cao2024maplm,yang2023lidar} explore LVLMs to address autonomous driving problems based on Nuscenes~\cite{caesar2020nuscenes} in roadside settings rather than the comprehension of city landscapes along with their spatial characteristics.
To bridge this gap, we propose an LVLM capable of understanding multiscale outdoor scenes, ranging from roadside environments to city landscapes.

\section{SVM-City}
\label{data_generation}
\begin{table}
\centering
  \caption{Comparison with scene understanding datasets.}
  \label{tab:comparison}
  \scalebox{0.75}{

\begin{tabular}{@{}c|c|cc|c|c|c@{}}
\toprule
\multirow{2}{*}{Data} &
  \multirow{2}{*}{Area} &
  \multicolumn{2}{c|}{Modality} &
  \multirow{2}{*}{Scale} &
  \multirow{2}{*}{Viewpoint} &
  \multirow{2}{*}{\begin{tabular}[c]{@{}c@{}}QA \\ Pairs\end{tabular}} \\ \cmidrule(lr){3-4}
 &
   &
  2D &
  3D &
   &
   &
   \\ \midrule
ScanRefer~\cite{chen2020scanrefer} &
  Indoor &
  \XSolidBrush &
  \Checkmark &
  Single  &
  Humanoid View &
  51k \\
Referit3d~\cite{achlioptas2020referit3d} &
  Indoor &
  \XSolidBrush &
  \Checkmark &
  Single &
  Humanoid View &
  125k \\
ScanQA~\cite{azuma2022scanqa} &
  Indoor &
  \XSolidBrush &
  \Checkmark &
  Single  &
  Humanoid View &
  41k \\
City-3DQA~\cite{sun20243d} &
  Outdoor &
  \XSolidBrush &
  \Checkmark &
  Single  &
  Low-altitude View &
  460k \\
NuscenesQA~\cite{qian2024nuscenes} &
  Outdoor &
  \Checkmark &
  \Checkmark &
  Single  &
  Terrestrial View &
  450k \\
EarthVQA~\cite{wang2024earthvqa} &
  Outdoor &
  \Checkmark &
  \XSolidBrush &
  Single  &
  High-altitude View &
  145k \\
KITTI360Pose~\cite{kolmet2022text2pos} &
  Outdoor &
  \XSolidBrush &
  \Checkmark &
  Single  &
  Terrestrial View &
  43k \\ \midrule
SVM-City (ours) &
  Outdoor &
  \Checkmark &
  \Checkmark &
  \begin{tabular}[c]{@{}c@{}}Multiple\\ Scales\end{tabular} &
  \begin{tabular}[c]{@{}c@{}}Terrestrial, Low-altitude,\\ High-altitude View\end{tabular} &
  567k \\ \bottomrule
\end{tabular}

}
\end{table}

\subsection{Data Generation}
In this section, we introduce the instruction tuning datasets from the multi\textbf{S}cale outdoor city-level scene based on multi\textbf{V}iew observation and multi\textbf{M}odal data, called \textbf{SVM-City}.
We split the data generation process into visual semantic collection, question template taxonomy, and data annotation.

\noindent \textbf{Visual Semantic Collection.} 
Given the large scale and complexity of outdoor urban scenes~\cite{tanner2022large}, we gather multidomain perception visual semantics, as summarized in Table~\ref{fig:pieline_annotation} (a). 
We realize the city scene understanding based on different observation scales (i.e. terrestrial, low-altitude, and high-altitude scales) \cite{macleod1999space}.
Terrestrial observations involve a single drive covering less than $1$ kilometer of urban streets~\cite{caesar2020nuscenes}. 
Low-altitude observations encompass community areas ranging from approximately $5$ to $20$ kilometers~\cite{yang2023urbanbis}. 
High-altitude observations cover entire metropolitan areas, typically spanning from a few hundred to several thousand kilometers~\cite{wang2loveda}.
The data can be divided into two categories: 3D and 2D.
Specifically, the Nuscenes dataset~\cite{caesar2020nuscenes} was acquired from vehicle-mounted sensors, while the LoveDA dataset~\cite{wang2loveda} consists of spaceborne RS imagery. 
The Earthexplorer dataset~\cite{EarthExp42:online} contains aerial orthophotos. Additionally, both the UrbanBIS~\cite{yang2023urbanbis} and SensatUrban~\cite{hu2022sensaturban} datasets are obtained from low-altitude drones.

\noindent \textbf{Question Template Taxonomy.} 
Based on the taxonomy of spatial questions proposed in cognition~\cite{hayward1995spatial}, we propose the following question templates for applications in outdoor scenes.

 $\bullet$ Localization. 
These questions aim to assess both the existence and spatial arrangement of objects within a city environment. 
For example, the question in this category might ask:\textit{"Where can municipal buildings be found in a city environment?"}

 $\bullet$ Measurement.
These questions pertain to providing information about the size, shape, and quantity of individual objects within an urban environment.
This category includes questions such as "\textit{How many buildings are in this city?}"

 $\bullet$ Functionality.
These questions aim to understand and infer the purpose, function, or affordance of objects within an outdoor city scene. 
For example, one might ask, "\textit{Which direction should I take to reach the art exhibition?}" 

$\bullet$ Logicality.
The purpose of questions is to establish the relative relationships between objects and scenes, which requires logical reasoning. For instance, consider the question: "\textit{Which car is closer to me, the blue one or the black one?}"

\begin{figure}
  \centering
  \includegraphics[width=0.78\linewidth]{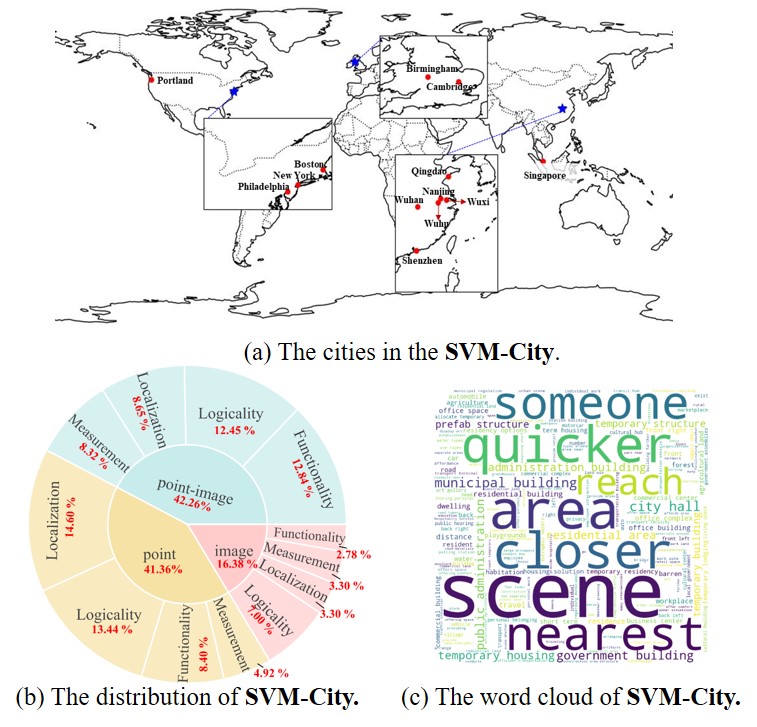}
  \caption{
  The statistics of SVM-City.
  }
  \label{fig:distribution}
\end{figure}

\noindent \textbf{Data Annotation.} 
We propose a comprehensive pipeline for constructing SVM-City, illustrated in Figure~\ref{fig:pieline_annotation}. 
This approach leverages multimodal outdoor scenes and predefined 2D or 3D segmentation methods (such as HRNet and B-Seg~\cite{wang2loveda,yang2023urbanbis}) to extract object sets through segmentation or bounding boxes. 
These are then used to generate spatial semantics using scene graphs automatically. 
These spatial semantics establish connections between objects and define their relationships. 
In addition, manually annotated attributes such as color (e.g., \textit{blue, yellow, green}), pose (e.g., \textit{moving, standing}), and functionality (e.g., \textit{residential area}, \textit{the location for shopping}) are incorporated into the spatial semantics. 
We further employ ChatGPT to automatically generate QA pairs based on the triples within the spatial semantics, utilizing predefined question templates to ensure language diversity and grammatical correctness.

\begin{figure*}
  \centering
  \includegraphics[width=0.75\linewidth]{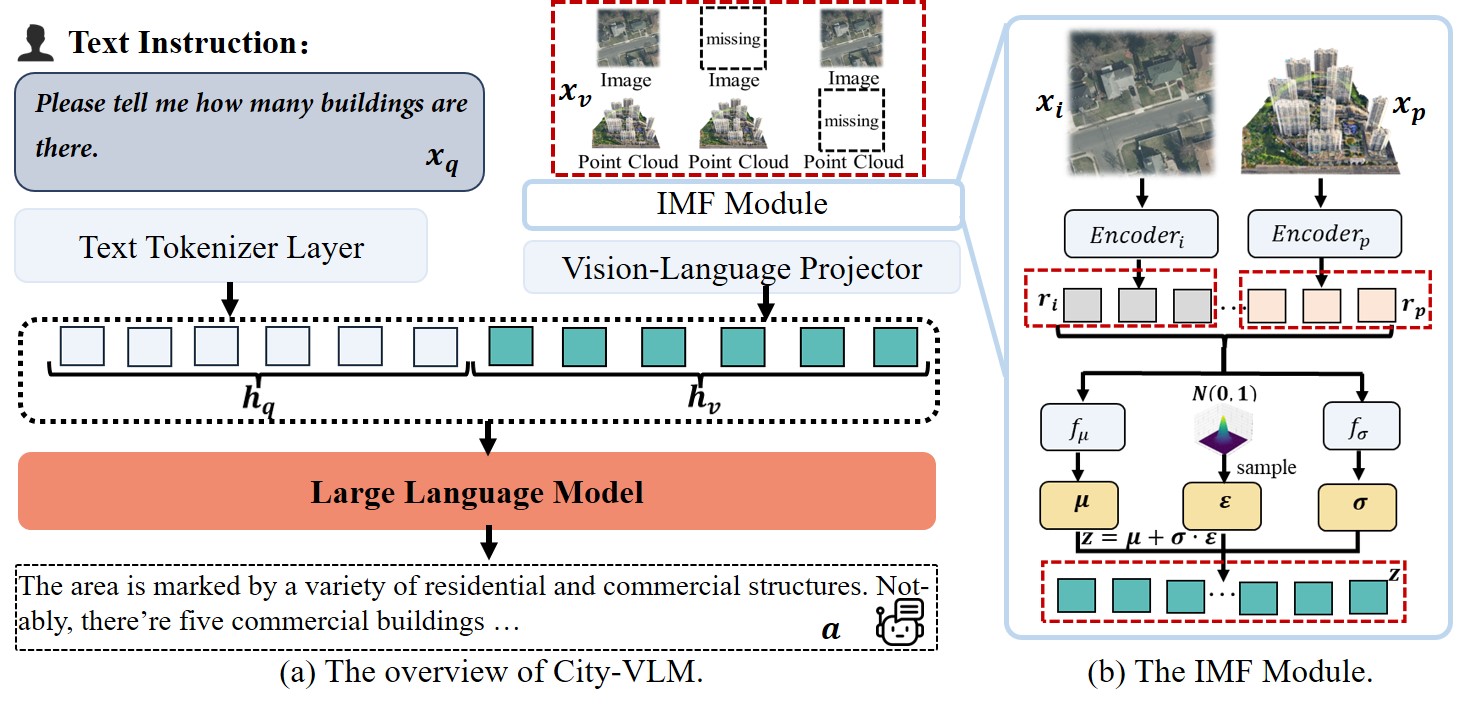}
  \caption{
  \spl{The architecture of City-VLM. The left (a) is the overview of City-VLM and the right (b) is the Incomplete Multimodal Fusion Module (IMF) Module.}
  }
  \label{fig:model}
\end{figure*}

\subsection{Data Statistics}
\label{app:statistics_for_sVM-city}

Figure~\ref{fig:distribution} presents the statistical distribution of cities in the SVM-City dataset. 
As shown in panel (a), the cities are grouped into three regions: North America, Western Europe, and East Asia. 
In North America, the dataset includes four cities: New York, Boston, Portland, and Philadelphia. 
In Western Europe, two cities are represented: Birmingham and Cambridge. 
Finally, in East Asia, there are seven cities: Qingdao, Nanjing, Wuhan, Wuxi, Wuhu, Shenzhen, and Singapore.
These cities represent some of the world's largest urban agglomerations and encompass many common characteristics shared among cities.
Figure \ref{fig:distribution}(b) illustrates the distribution of questions according to their corresponding templates. 
The SVM-City dataset is categorized into three visual modalities: point, image, and point-image (a combination of both point and image). These modalities constitute $41.36 \%$, $16.38 \%$, and $42.26 \%$ of the dataset, respectively. Each modality encompasses four types of questions, as detailed in Section~\ref{data_generation}: Localization, Measurement, Functionality, and Logicality. The distribution of question types is as follows: Localization accounts for $26.55 \%$, Measurement for $16.54 \%$, Functionality for $24.02 \%$, and Logicality for $32.89 \%$.
Figure \ref{fig:distribution}(C) shows the most frequent words in SVM-City. 
SVM-City includes a relatively rich vocabulary and a variety of phrases, due to the polishing and grammar correction provided by ChatGPT.

We compare SVM-City with other scene understanding datasets in Table~\ref{tab:comparison}. Existing scene understanding datasets for indoor and outdoor environments typically focus on a single scale. 
Indoor datasets are generally captured from a humanoid perspective, while outdoor datasets are observed from terrestrial, high-altitude, or low-altitude viewpoints. 
\spl{
In contrast, Our work focuses on city-level urban scene understanding, which fundamentally differs from the roadside scene understanding targeted by autonomous driving datasets (e.g., KITTI360Pose, NuscenesQA).
Specifically, Autonomous driving datasets primarily capture roadside scenes (e.g., vehicles, pedestrians, traffic signs) within a limited spatial range, relying on vehicle-mounted sensors (e.g., LiDAR, RGB cameras)..
Our dataset covers entire urban areas, including buildings, functional zones, and city structures, enabling a holistic understanding of urban environments.
Our dataset incorporates not only vehicle-mounted sensors, but also multi-source data including drone, aerial plane and satellite, providing richer information for city-level analysis.
}

\section{City-VLM}

\subsection{Network Overview}
\spl{
We introduce City-VLM, a vision-language model trained on the SVM-City dataset, as shown in Figure~\ref{fig:model} (a). 
The model takes urban visual data $\mathbf{x_v}$ with language queries $\mathbf{x_q}$ as input and generates the final answers $\mathbf{a}$.
First, the visual input $\mathbf{x_v}$ is encoded into a probabilistic visual embedding $\mathbf{z}$ through an Incomplete Multimodal Fusion Module (IMF Module),
\begin{equation}
\scalebox{0.9}{$
\begin{aligned} 
\mathbf{z} = \text{IMF}(\mathbf{x_v}).
\end{aligned}
$}
\end{equation}
The embedding $\mathbf{z}$ is then projected into a sequence of vision-language tokens $\mathbf{h_v}$ via a vision-language projector. 
Concurrently, the text input $\mathbf{x_q}$ is tokenized into text tokens $\mathbf{h_q}$ using a standard text tokenizer. 
The vision tokens $\mathbf{h_v}$ and text tokens $\mathbf{h_q}$ are concatenated to form the input sequence for the large language model (LLM), which autoregressively generates the output sequence $\mathbf{a}$:
\begin{equation}
\scalebox{0.9}{$
\begin{aligned} 
\text{P}(\mathbf{a}|\mathbf{h_v},\mathbf{h_q}) =\prod_{j=1}^{L}\text{P}(a_j|\mathbf{h_v},\mathbf{h_q},a_{<j}),
\end{aligned}
$}
\end{equation}
where $L$ denotes the length of $\mathbf{a}$ and $a_j$ is the output answer token of $\mathbf{a}$.

Let the tuple $\mathbf{x_v}$ = $(\mathbf{x_i}, \mathbf{x_p})$ represent the visual input modalities in an urban environment, where $\mathbf{x_i}$ denotes the image modality and $\mathbf{x_p}$ represents the point cloud modality.
In the complete case, all modalities are observed and easily fused to feed the downstream tasks. 
However, in practical outdoor environments, certain modalities may be missing, requiring specialized processing to achieve effective fusion.
}

\subsection{Incomplete Multimodal Fusion Module}

\spl{
We introduce incomplete multimodal learning into the Incomplete Multimodal Fusion Module (IMF Module) to model both the mean and variance of missing modalities as learnable parameters as shown in Figure~\ref{fig:model} (b). 
When either the 2D image $\mathbf{x_{i}}$ or the 3D point cloud $\mathbf{x_{p}}$ is absent, we pad the missing input with the zero tensor to ensure all input modalities have consistent dimensions and formats following the existing incomplete multimodal researches~\cite{pan2021disease, wei2024robust}.
The input image $\mathbf{x_{i}}$ and point cloud $\mathbf{x_{p}}$ have the dedicated encoders to extract their feature representation $\mathbf{r_i}$ and $\mathbf{r_p}$, as described by the equation:
\begin{equation}
\scalebox{0.9}{$
\begin{aligned} 
\mathbf{r_{i}} = Encoder_{i}(\mathbf{x_{i}}), \\
\mathbf{r_{p}} = Encoder_{p}(\mathbf{x_{p}}).
\end{aligned}
$}
\end{equation}
We concatenate $\mathbf{r_i}$ and $\mathbf{r_p}$ to obtain the the modality feature representation $\mathbf{r_z}$.

To make the model robust to missing modalities,  $\mathbf{r_z}$ is modeled as a probabilistic distribution based on the VAE method~\cite{kingma2013auto}. 
Specifically, we build the probabilistic embeddings $\mathbf{r_z} \sim p(\mathbf{r_z} \mid \mathbf{r_i}, \mathbf{r_p})$ and adopt the Gaussian distribution,  

\begin{equation}
\scalebox{0.9}{$
\begin{aligned} 
p(\mathbf{r_z} \mid \mathbf{r_i}, \mathbf{r_p})=\mathcal{N} (\mathbf{r_z};\boldsymbol{\mu},\boldsymbol{\sigma}^2),
\end{aligned}
$}
\end{equation}
where $\boldsymbol{\mu}$ and $\boldsymbol{\sigma}$  are the mean and variance of the distribution, calculated from the modality-specific feature representations,
\begin{equation}
\scalebox{0.9}{$
\begin{aligned} 
\boldsymbol{\mu} = f_{\mu}(\mathbf{r_{z}}),\quad \text{log}(\boldsymbol{\sigma})=f_{\sigma}(\mathbf{r_{z}}),
\end{aligned}
$}
\end{equation}
where $f_{\mu}(\cdot)$ and $f_{\sigma}(\cdot)$ are the functions used to estimate $\boldsymbol{\mu}$ and $\boldsymbol{\sigma}$.

To allow backpropagation through the sampling process, we apply the reparameterization trick~\cite{kingma2013auto}. 
We sample from the Gaussian distribution by adding noise $\boldsymbol{\varepsilon}$ sampled from $\mathcal{N}(0,\mathbf{I})$ to the mean representation:
\begin{equation}
\scalebox{0.9}{$
\begin{aligned} 
\mathbf{z} = \boldsymbol{\mu}+\boldsymbol{\varepsilon} \cdot \boldsymbol{\sigma}, \quad \boldsymbol{\varepsilon} \sim \mathcal{N}(0,1),
\end{aligned}
$}
\end{equation}
where $\mathbf{z}$ is the representation used for training, while $\boldsymbol{\mu}$ is the representation used for inference.
Inspired by previous probabilistic embedding methods~\cite{chang2020data}, we introduce a regularization term in the optimization process by explicitly constraining $\mathcal{N} (\mathbf{r_z};\boldsymbol{\mu},\boldsymbol{\sigma}^2)$ to be close to a standard normal distribution $\mathcal{N}(0, \mathbf{I})$ with the KL divergence,
\begin{equation}
\scalebox{0.9}{$
\begin{aligned} 
L_{kl} &= KL[\mathcal{N} (\mathbf{r_z};\boldsymbol{\mu},\boldsymbol{\sigma}^2) || \mathcal{N}(0, \mathbf{I})]\\
&=-\frac{1}{2} (1+log (\boldsymbol{\sigma}^2)-\boldsymbol{\mu}^2-\boldsymbol{\sigma}^2 ).
\end{aligned}
$}
\end{equation}
}

\section{Experiments}

\subsection{Implementation Details}
\noindent \textbf{Training Details}.
We employ a cross-modal alignment encoder, utilizing the Uni3D-L~\cite{zhou2023uni3d} as the 3D encoder, EVA-CLIP-E~\cite{radford2021learning} as the 2D encoder, and Vicuna-7B~\cite{chiang2023vicuna} as the large language model. 
\spl{
we utilize the vision encoder in the fixed resolution following the existing works in LVLMs~\cite{kuckreja2024geochat,bazi2024rs}.
This design follows the existing large vision language model design and ensures stable positional encoding in Transformer architectures. 
Besides, our downstream task focuses on question answering, and the task demonstrates robustness to pixel-level deviations.
}
Our City-VLM is trained on the SVM-City dataset.
Experiments are implemented with CUDA $11.8$ and PyTorch $2.0.1$ and run on $8$ NVIDIA RTX A6000. 
We employ the Adam optimizer with weight decay $5e^{-4}$, a learning rate of $1e^{-3}$, and a batch size of $4$ on each device during the training stage in the LoRA~\cite{hulora} setting.

\noindent \textbf{Evaluation Tasks}.
We evaluate our method on three QA tasks covering high-altitude, low-altitude and terrestrial view, EarthVQA~\cite{wang2024earthvqa}, City-3DQA~\cite{sun20243d} and Nuscenes-QA~\cite{qian2024nuscenes}. 

$\bullet$ \textbf{EarthVQA} dataset in test comprises $1,809$ high-resolution remote sensing 2D images with $63,225$ QA pairs. 
The questions in EarthVQA are categorized into six types: basic judgment (Bas Ju), reasoning-based judgment (Rel Ju), basic counting (Bas Co), reasoning-based counting (Rel Co), object situation analysis (Obj An), and comprehensive analysis (Com An).

$\bullet$ \textbf{City-3DQA} contains $2.5$ billion 3D point clouds collected via drone and supports two modes of evaluation: sentence-wise and city-wise. 
Each mode's test set includes $78$k QA pairs, divided into single-hop and multi-hop questions. 
Single-hop questions are those that can be answered using direct inference, while multi-hop questions require integrating multiple pieces of information through a series of reasoning steps.
The sentence-wise mode has $34$k single-hop and $44$k multi-hop questions, while the city-wise mode comprises $37$k single-hop and $41$k multi-hop questions.

$\bullet$ \textbf{Nuscenes-QA} dataset is the QA dataset in the autonomous-driving setting. The test set consists of $83$k QA pairs, accompanied by $390$k LiDAR point clouds and $1.4$ million camera images, all captured from a vehicle-mounted system. 
The questions in the dataset are categorized into five types based on their query format: Exist, Count, Object, Status, and Comparison. 

\begin{table*}
\centering
  \caption{\spl{Comparison with other VQA methods on EarthVQA. Seg. denotes the model using deep semantic segmentation. OA means the overall accuracy.}}
  \label{tab:earthvqa}
  \scalebox{0.88}{
 \begin{tabular}{@{}cc|c|cccccc|c@{}}
\toprule
\multicolumn{2}{c|}{\multirow{2}{*}{Method}}                                                                             & \multirow{2}{*}{Seg.} & \multicolumn{6}{c|}{$\uparrow$ Accuracy (\%)}                                                       & \multirow{2}{*}{$\uparrow$ OA (\%)} \\ \cmidrule(lr){4-9}
\multicolumn{2}{c|}{}                                                                                                    &                       & Bas Ju         & Rel Ju         & Bas Co         & Rel Co         & Obj An         & Com An         &                                     \\ \midrule
\multicolumn{1}{c|}{\multirow{9}{*}{Specialist Models}} & SAN~\cite{yang2016stacked}               & $\times$              & 87.59          & 81.79          & 76.26          & 59.23          & 55.00          & 43.25          & 75.66                               \\
\multicolumn{1}{c|}{}                                   & MAC~\cite{hudson2018compositional}       & $\times$              & 82.89          & 79.46          & 72.53          & 55.86          & 46.32          & 40.50          & 71.98                               \\
\multicolumn{1}{c|}{}                                   & BUTD~\cite{anderson2018bottom}           & $\checkmark$          & 90.01          & 82.02          & 77.16          & 60.95          & 56.29          & 42.29          & 76.49                               \\
\multicolumn{1}{c|}{}                                   & BAN~\cite{kim2018bilinear}               & $\checkmark$          & 89.81          & 81.87          & 77.58          & 63.71          & 55.67          & 45.06          & 76.74                               \\
\multicolumn{1}{c|}{}                                   & MCAN~\cite{yu2019deep}                   & $\checkmark$          & 89.65          & 81.65          & 79.83          & 63.16          & 57.28          & 43.71          & 77.01                               \\
\multicolumn{1}{c|}{}                                   & D-VQA~\cite{wen2021debiased}             & $\checkmark$          & 89.73          & 82.12          & 77.38          & 63.99          & 55.14          & 43.20          & 76.59                               \\
\multicolumn{1}{c|}{}                                   & RSVQA~\cite{lobry2020rsvqa}              & $\times$              & 82.43          & 79.34          & 70.68          & 55.53          & 42.45          & 35.46          & 70.70                               \\
\multicolumn{1}{c|}{}                                   & RSIVQA~\cite{zheng2021mutual}            & $\times$              & 85.32          & 80.44          & 75.01          & 56.63          & 51.55          & 39.25          & 73.70                               \\
\multicolumn{1}{c|}{}                                   & SOBA~\cite{wang2024earthvqa}             & $\checkmark$          & 89.63          & 82.64          & 80.17          & \textbf{67.86} & 61.40          & 49.30          & 78.14                               \\ \midrule
\multicolumn{1}{c|}{\multirow{5}{*}{LVLMs}}             & BLIP-2~\cite{li2023blip}                 & $\times$              & 88.13          & 81.92          & 70.26          & 58.58          & 42.72          & 28.34          & 71.07                               \\
\multicolumn{1}{c|}{}                                   & Instruct-BLIP~\cite{instructblip2023dai} & $\times$              & 89.67          & 79.69          & 76.96          & 63.34          & 59.72          & 45.68          & 75.25                               \\
\multicolumn{1}{c|}{}                                   & City-VLM w/ Attention (ours)                                   & $\times$              & 90.47          & 81.09          & 78.68          & 63.60          & 65.35          & 44.24          & 76.91                               \\
\multicolumn{1}{c|}{}                                   & City-VLM w/ MLP (ours)                                         & $\times$              & 91.35          & 81.30          & 80.01          & 64.49          & 64.89          & 43.96          & 77.40                               \\
\multicolumn{1}{c|}{}                                   & City-VLM w/ IMF (ours)                                         & $\times$              & \textbf{91.42} & \textbf{82.71} & \textbf{80.23} & 65.21          & \textbf{66.83} & \textbf{49.97} & \textbf{78.84}                      \\ \bottomrule
\end{tabular}
 
}
\end{table*}

\noindent \textbf{Evaluation Metrics}.
Previous tasks commonly use classification accuracy as the evaluation metric. 
However, this approach is not suitable for our auto-regressive model. 
Drawing inspiration from LLaVA series work~\cite{liu2024visual,liu2024improved}, we utilize GPT-4 to assess the quality of the generated responses and we the evaluation code in this link~\footnote{~\url{https://github.com/haotian-liu/LLaVA/tree/main/llava/eval}}.
We use the following prompt in LLaVA work:
\begin{tcolorbox}[
left=3mm, right=3mm, top=0.001mm, bottom=0.001mm,fontupper=\small,  
]
\textit{  \hspace*{0.7em} Analyze two sentences and determine if they're referring to the same general object or concept, focusing on the type of object, not attributes such as color, size, or shape. Respond with `T' if they refer to the same thing and `F' if not. Also, provide a brief rationale for your judgment.}

\textit{\hspace*{0.7em} Now, let's analyze the following:}

\textit{\hspace*{0.7em} Input: 1. \{ground\_truth\} 2. \{model\_output\}}

\textit{\hspace*{0.7em} Output:}
\end{tcolorbox}




Specifically, we construct triplets composed of the generated responses from our model, the corresponding ground-truth language answers, and the questions. 
These triplets are then input to a judge (i.e., GPT-4 in text-only mode), which evaluates whether the generated responses convey the same meaning as the ground-truth answers, based on the given questions.
The final accuracy is calculated based on the judge’s assessments.

\spl{
\noindent \textbf{Ablation Study}.
To evaluate the effectiveness of the IMF Module in City-VLM (City w/ IMF), we design two ablation models. 
In ablation models, the IMF Module is replaced with an MLP module and a cross-attention module, referred to as \textbf{City-VLM w/ MLP} and \textbf{City-VLM w/ Attention}, respectively.
These methods represent the most widely used techniques for mapping and merging module in LVLMs~\cite{liu2024improved,li2023blip,instructblip2023dai}.
These models follow existing incomplete multimodal researches to handle missing input by padding zeros~\cite{pan2021disease, wei2024robust}. 
}

\subsection{EarthVQA Comparative Experiments}

\noindent \textbf{Baselines.}
We classify the public baseline methods has been applied in EarthVQA~\cite{wang2024earthvqa} into two categories: specialist models and LVLMs. 
Specialist models are designed for the remote sense QA tasks, including SAN~\cite{yang2016stacked}, MAC~\cite{hudson2018compositional}, BUTD~\cite{anderson2018bottom}, BAN~\cite{kim2018bilinear}, MCAN~\cite{yu2019deep}, D-VQA~\cite{wen2021debiased}, RSVQA~\cite{lobry2020rsvqa}, RSIVQA~\cite{zheng2021mutual} and SOBA~\cite{wang2024earthvqa}.
Besides, BUTD, BAN, MCAN, D-VQA and SOBA take deep semantic segmentation as the auxiliary information for the remote sense interpretation.
The baseline LVLM models including Instruct-BLIP and BLIP-2 are fine-tuned on the EarthVQA following the existing baseline setting~\cite{wang2024earthvqa}.

\noindent \textbf{Quantitative results.}
We evaluate the performance of models on high-altitude view outdoor scenes using the EarthVQA dataset~\footnote{\url{https://www.codabench.org/competitions/2922/}}. 
The comparison of the results is presented in Table~\ref{tab:earthvqa}.
Our proposed City-VLM model (City-VLM w/ IMF) establishes better performance on the test set, achieving a score of $78.84 \%$. 
This marks an improvement of $1.83 \%$, $0.7 \%$ over the best specialist models ($78.14 \%$ $\to$ $78.84 \%$).
Specifically, while the current state-of-the-art model, SOBA, uses deep semantic segmentation as supplementary information to retrieve answers within a constrained response space through multilayer perceptrons (MLPs), it faces challenges when the answer extends beyond the pre-defined scope. 
Our model demonstrates superior performance without relying on semantic segmentation features and is capable of handling a broader range of potential answers.

Other LVLMs, such as BLIP-2 and Instruct-BLIP, achieve scores of $71.07 \%$ and $75.25 \%$, respectively, which are lower compared to specialist models (e.g., SOBA~\cite{wang2024earthvqa}) ($78.14 \%$). 
However, our model City-VLM achieves $3.59 \%$ accuracy over the top LVLMs ($75.25 \% \to 78.84 \%$).
We attribute this performance to the training data from the SVM-City dataset, which incorporates additional high-altitude data, such as aerial orthophotos. 
This enriched dataset has significantly enhanced the model's ability to interpret remote sensing imagery.

\subsection{City-3DQA Comparative Experiments}

\begin{table*}
\centering
  \caption{\spl{Comparsion with other 3D QA and LLM methods on City-3DQA. Sentence-wise and City-wise denote different sets of City-3DQA.}}
  \label{tab:city3dqa}
  \scalebox{0.88}{

\begin{tabular}{@{}cc|ccc|ccc@{}}
\toprule
\multicolumn{2}{c|}{\multirow{2}{*}{Methods}}                                                                                                & \multicolumn{3}{c|}{$\uparrow$ Sentence-wise (\%)} & \multicolumn{3}{c}{$\uparrow$ City-wise (\%)}    \\ \cmidrule(l){3-8} 
\multicolumn{2}{c|}{}                                                                                                                        & Single-hop      & Multi-hop       & All            & Single-hop     & Multi-hop      & All            \\ \midrule
\multicolumn{1}{c|}{\multirow{5}{*}{\begin{tabular}[c]{@{}c@{}}Specialist\\ Models\end{tabular}}}       & ScanQA~\cite{azuma2022scanqa}      & 76.42           & 28.31           & 49.28          & 64.84          & 27.03          & 47.33          \\
\multicolumn{1}{c|}{}                                                                                   & CLIP-Guided~\cite{parelli2023clip} & 74.54           & 33.73           & 51.55          & 63.05          & 32.41          & 46.94          \\
\multicolumn{1}{c|}{}                                                                                   & 3D-VLP~\cite{jin2023context}       & 72.78           & 35.54           & 51.72          & 64.03          & 34.95          & 48.74          \\
\multicolumn{1}{c|}{}                                                                                   & 3D-VisTA~\cite{zhu20233d}          & 79.23           & 44.67           & 59.63          & 71.28          & 43.87          & 56.74          \\
\multicolumn{1}{c|}{}                                                                                   & Sg-CityU~\cite{sun20243d}          & 80.95           & 50.75           & 63.94          & 78.46          & 50.50          & 63.76          \\ \midrule
\multicolumn{1}{c|}{\multirow{7}{*}{\begin{tabular}[c]{@{}c@{}}Large\\ Language\\ Models\end{tabular}}} & Qwen-VL~\cite{Qwen-VL}             & 30.53           & 9.76            & 18.81          & 30.79          & 9.78           & 19.75          \\
\multicolumn{1}{c|}{}                                                                                   & LLaVA~\cite{liu2024visual}         & 33.93           & 10.33           & 20.60          & 32.56          & 9.84           & 20.56          \\
\multicolumn{1}{c|}{}                                                                                   & Qwen~\cite{qwen}                   & 55.25           & 11.21           & 30.35          & 55.40          & 12.59          & 31.31          \\
\multicolumn{1}{c|}{}                                                                                   & Llama-2~\cite{touvron2023llama}    & 60.51           & 20.00           & 37.66          & 60.03          & 18.82          & 38.37          \\
\multicolumn{1}{c|}{}                                                                                   & City-VLM w/ Attention (ours)       & 80.66           & 52.53           & 64.36          & 77.42          & 50.63          & 62.80          \\
\multicolumn{1}{c|}{}                                                                                   & City-VLM w/ MLP (ours)             & 80.47           & 52.92           & 64.51          & 77.55          & 51.90          & 63.55          \\
\multicolumn{1}{c|}{}                                                                                   & City-VLM w/ IMF (ours)              & \textbf{81.74}  & \textbf{56.80}  & \textbf{67.30} & \textbf{78.84} & \textbf{52.26} & \textbf{64.70} \\ \bottomrule
\end{tabular}

}
\end{table*}

\noindent \textbf{Baselines.}
Several public are baseline models for our experiments, including ScanQA, CLIP-Guided, 3D-VLP, 3D-VisTA, and the SOTA model Sg-CityU following City-3DQA~\cite{sun20243d}. Notably, ScanQA, CLIP-Guided, 3D-VLP, and 3D-VisTA process point cloud data as their primary input, while Sg-CityU incorporates both point cloud data and a scene graph that encodes spatial semantics following the existing baseline setting~\cite{sun20243d}.
Furthermore, following the  City-3DQA~\cite{sun20243d}, several baselines LLM are divided into two types: LVLM utilizing 2D images (Qwen-vl and LLaVA) and LLM (Qwen and Llama-2) utilizing scene graphs as input. For the former, we convert the input point clouds into 2D images. 
For the latter, we construct the scene graph from space semantic of each city scene and we organize these scene graphs in language.

\noindent \textbf{Quantitative results.}
To evaluate model performance on low-altitude view outdoor scenes, we employ the City-3DQA dataset and the comparison results are shown in Table~\ref{tab:city3dqa}.
Our model (City-VLM w/ IMF) achieves $67.30 \%$ and $64.70 \%$ in sentence-wise and city-wise set of City-3DQA, over existing model Sg-CityU~\cite{sun20243d} $3.36 \% (63.94 \% \to 67.30)$ and $0.94 \%  (63.76 \% \to 64.70 \%)$ respectively.
Similar to SOBA used in EarthVQA, Sg-CityU also employs MLPs to select answers from a predefined answer space. 
In contrast, City-VLM can access a broader range of possible answers.
Specifically, our model demonstrates consistent improvements over Sg-CityU in both sentence-wise and city-wise evaluation metrics.
These results indicate that our approach consistently outperforms the baseline across different types of QA tasks, particularly in multi-hop reasoning. 
We attribute these improvements to the enhanced reasoning capabilities embedded in the large language model employed by City-VLM~\cite{kim2024towards}.

The general LVLMs, including Qwen-VL~\cite{Qwen-VL} and LLaVA~\cite{liu2024visual}, project 3D point clouds into 2D images and answer questions based on the projected images.
The current leading general-purpose vision-language models achieve accuracy rates of $20.60 \%$ for sentence-wise City-3DQA and $20.56 \%$ for city-wise City-3DQA. 
In contrast, our model, City-VLM (City-VLM w/ IMF), surpasses these methods, achieving over $40 \%$ accuracy. 
This significant improvement indicates that existing general vision-language models struggle with low-altitude scene understanding.


\subsection{Nuscenes-QA Comparative Experiments}

\begin{table*}
\centering
  \caption{\spl{Comparison with other VQA methods on Nuscenes-QA. C and L denote camera and lidar.}}
  \label{tab:nuscenes}
  \scalebox{0.88}{

\begin{tabular}{@{}cc|c|ccccc|c@{}}
\toprule
\multicolumn{2}{c|}{Methods}                                                                                                                      & Modality & Exist         & Count         & Object        & Status        & Comparison    & $\uparrow$ Acc (\%) \\ \midrule
\multicolumn{1}{c|}{\multirow{6}{*}{\begin{tabular}[c]{@{}c@{}}Specialist\\ Models\end{tabular}}} & BEVDet+BUTD~\cite{huang2021bevdet}            & C        & 83.7          & 20.9          & 48.4          & 52.0          & 67.7          & 57.0                \\
\multicolumn{1}{c|}{}                                                                             & CenterPoint+BUTD~\cite{yin2021center}         & L        & 84.1          & 21.3          & 49.2          & 55.9          & 69.2          & 58.1                \\
\multicolumn{1}{c|}{}                                                                             & MSMDFusion+BUTD~\cite{jiao2023msmdfusion}     & C+L      & 85.1          & \textbf{23.2} & 52.3          & 59.5          & 68.5          & 59.8                \\ \cmidrule(l){2-9} 
\multicolumn{1}{c|}{}                                                                             & BEVDet+MCAN~\cite{huang2021bevdet}            & C        & 84.2          & 20.4          & 51.2          & 54.7          & 67.4          & 57.9                \\
\multicolumn{1}{c|}{}                                                                             & CenterPoint+MCAN~\cite{yin2021center}         & L        & 84.8          & 20.8          & 52.3          & 59.8          & 70.0          & 59.5                \\
\multicolumn{1}{c|}{}                                                                             & MSMDFusion+MCAN~\cite{jiao2023msmdfusion}     & C+L      & 85.4          & 22.2          & 54.3          & 60.6          & 69.7          & 60.4                \\ \midrule
\multicolumn{1}{c|}{\multirow{11}{*}{LVLMs}}                                                      & LLaVA~\cite{liu2024visual}                    & C        & 73.8          & 14.6           & 37.9           & 45.9           & 53.3          & 47.4                \\
\multicolumn{1}{c|}{}                                                                             & LidarLLM~\cite{yang2023lidar}                 & L        & 74.5          & 15.0          & 37.8          & 45.9          & 57.8          & 48.6                \\ \cmidrule(l){2-9} 
\multicolumn{1}{c|}{}                                                                             & \multirow{3}{*}{City-VLM w/ Attention (ours)} & C        & 81.5          & 18.5          & 51.0          & 55.7          & 67.9          & 57.1                \\
\multicolumn{1}{c|}{}                                                                             &                                               & L        & 83.2          & 18.1          & 53.4          & 55.9          & 68.8          & 58.2                \\
\multicolumn{1}{c|}{}                                                                             &                                               & C+L      & 85.3          & 18.7          & 53.8          & 57.2          & 69.2          & 59.2                \\ \cmidrule(l){2-9} 
\multicolumn{1}{c|}{}                                                                             & \multirow{3}{*}{City-VLM w/ MLP (ours)}       & C        & 82.7          & 18.9          & 51.4          & 54.7          & 68.0          & 57.4                \\
\multicolumn{1}{c|}{}                                                                             &                                               & L        & 83.7          & 17.6          & 52.7          & 56.8          & 69.3          & 58.3                \\
\multicolumn{1}{c|}{}                                                                             &                                               & C+L      & 84.1          & 19.4          & 54.8          & 58.5          & 69.5          & 59.4                \\ \cmidrule(l){2-9} 
\multicolumn{1}{c|}{}                                                                             & \multirow{3}{*}{City-VLM w/ IMF (ours)}        & C        & 83.3          & 19.1          & 52.1          & 56.2          & 68.4          & 58.1                \\
\multicolumn{1}{c|}{}                                                                             &                                               & L        & 85.6          & 18.3          & 54.0          & 59.1          & 70.4          & 59.7                \\
\multicolumn{1}{c|}{}                                                                             &                                               & C+L      & \textbf{87.6} & 20.0          & \textbf{55.6} & \textbf{62.3} & \textbf{71.7} & \textbf{61.6}       \\ \bottomrule
\end{tabular}

}
\end{table*}

\noindent \textbf{Baselines.}
The current models utilize vehicle 2D camera images (C) and 3D lidar points (L) as input and we follow the existing baseline setting~\cite{qian2024nuscenes}.
The other baseline models can be divided into two parts, using BUTD~\cite{anderson2018bottom} and MCAN~\cite{yu2019deep} as fusion layer with pre-trained 2D or 3D object detection backbone, which is the general approach in the autonomous driving settings.
BEVDet~\cite{huang2021bevdet} for camera-only setting, which encodes the perspective-view features to detect the object bounding boxes. 
For the LiDAR-only setting, CenterPoint~\cite{yin2021center} introduces a center-based object keypoint detector. 
For the multi-modal model, MSMDFusion~\cite{jiao2023msmdfusion} leverages depth information and fine-grained cross-modal interactions between the LiDAR and camera for 3D object detection.
In LVLMs, we choose the public baseline models LLaVA~\cite{liu2024visual} and LidarLLM~\cite{yang2023lidar} which are fine-tuned on the 2D and 3D data subset respectively. 

\noindent \textbf{Quantitative results.}
To evaluate model performance on 3D and 2D city scenes, we employ the Nuscenes-QA dataset as a benchmark and the comparison results are shown in Table~\ref{tab:nuscenes}.
When utilizing only camera images, BEVDet+MCAN achieves an accuracy of $57.9 \%$. 
Our proposed model enhances this performance, achieving an improved accuracy of $58.1 \%$, representing an increase of $0.2 \%$. 
Similarly, when using only lidar points, CenterPoint+MCAN reaches an accuracy of $59.5 \%$. Our model demonstrates an improvement here as well, achieving an accuracy of $59.7 \%$, also representing a $0.2 \%$ increase.
City-VLM, which incorporates both image and point cloud data, achieves an accuracy of $61.6 \%$. This represents an improvement of $1.2 \%$ over the current SOTA model, MSMDFusion+MCAN (C + L), which reaches an accuracy of $60.4 \%$.
Specifically, our model outperforms existing state-of-the-art (SOTA) models across four question types: Existence, Object, Status, and Comparison, achieving improvements of $2.2 \% (85.4 \% \to 87.6 \%)$, $2.3 \% (54.3 \% \to 55.6 \%)$, $1.7 \% (60.6 \% \to 62.3 \%)$, and $2.0 \% (69.7 \% \to 71.7 \%)$, respectively.
Our model encounters difficulties with counting-related questions, which we attribute to the limitations of autoregressive methods in accurately handling counting tasks.

\subsection{Case Study}
\begin{figure}
  \centering
  \includegraphics[width=0.9\linewidth]{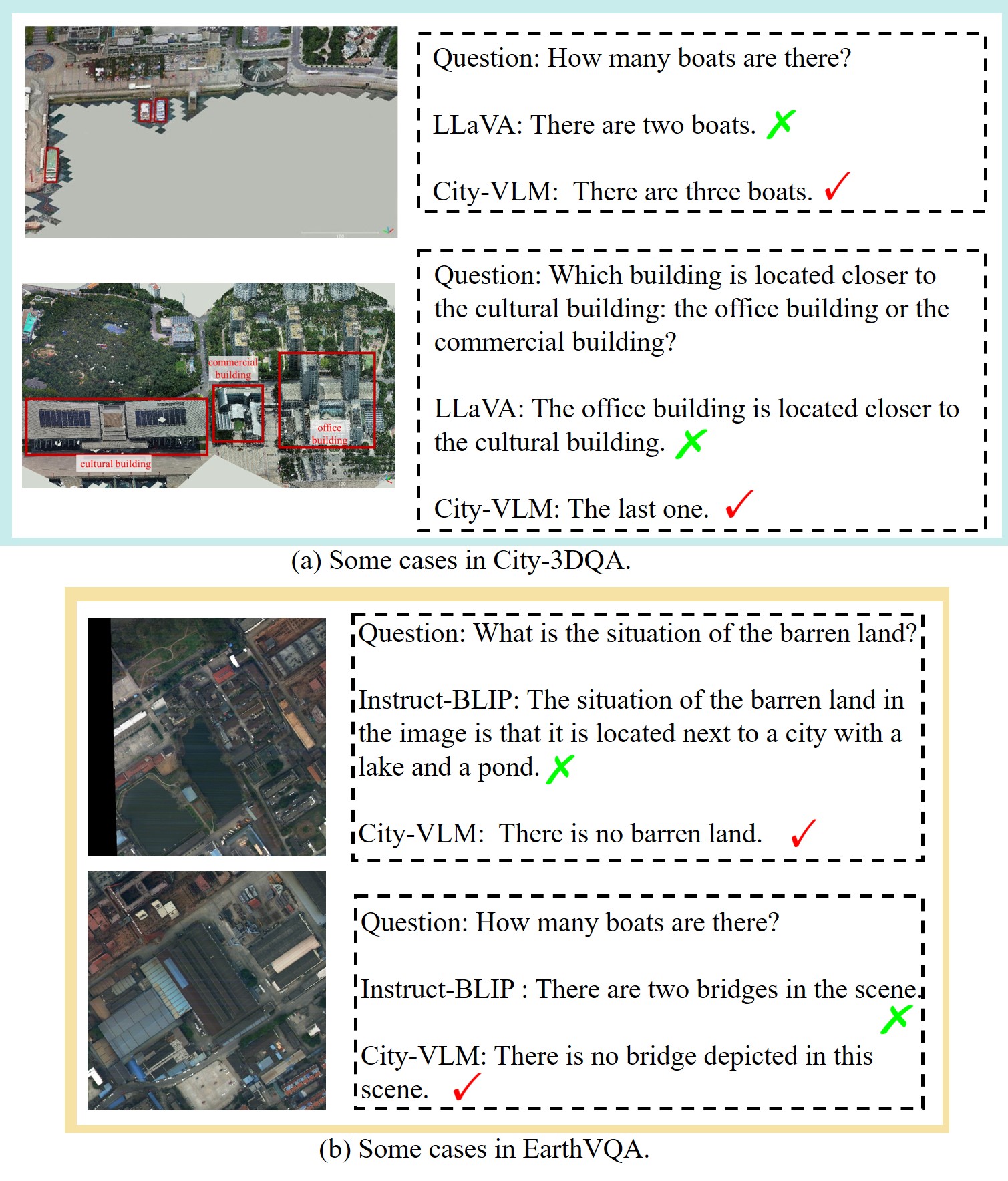}
  \caption{
  In this case studies, we compare the performance of existing LVLMs and City-VLM models.
  }
  \label{fig:case_study}
\end{figure}

Our model demonstrates superior performance compared to existing models when using equivalent modalities. 
Unlike most existing approaches, which rely heavily on pre-trained 2D or 3D object detection modules, our model eliminates the need for such components. 
These pre-trained detection modules typically require extensive manual annotation, leading to increased time and labor costs.
In contrast, our automated labeling process for the pre-training dataset significantly reduces these expenses. 
Our model achieves higher accuracy with a more cost-effective solution for data acquisition compared to current methods.

\subsection{Ablation Study}

\spl{
We conduct an ablation study to assess the effectiveness of the Incomplete Multimodal Fusion Module (IMF) Module, with results presented for different datasets in Tables~\ref{tab:earthvqa}, ~\ref{tab:city3dqa}, and ~\ref{tab:nuscenes}.
In the EarthVQA dataset, the models City-VLM w/ MLP and City-VLM w/ Attention achieve accuracies of $77.40 \%$ and $76.91 \%$, respectively. These are $1.44 \%$ and $1.93 \%$ lower than the accuracy of the City-VLM w/ IMF model, which achieves $78.84 \%$.
For the City-3DQA dataset, the City-VLM w/ MLP model achieves accuracy scores of $64.51 \%$ (sentence-wise) and $63.55 \%$ (city-wise), while the City-VLM w/ Attention model achieves scores of $64.36 \%$ (sentence-wise) and $62.80 \%$ (city-wise). Both models perform lower than their respective counterparts with the IMF Module. Specifically, the IMF Module improves sentence-wise accuracy by $2.79 \%$ (from $64.51 \%$ to $67.30 \%$) and city-wise accuracy by $1.15 \%$ (from $63.55 \%$ to $64.70 \%$).
In the Nuscenes-QA dataset, the City-VLM w/ MLP model, using both image and point-cloud inputs, achieves an accuracy of $59.4 \%$, which is $2.2 \%$ lower than the accuracy of the model with the IMF Module ($61.6 \%$). Similarly, the City-VLM w/ Attention model, also using the image and point-cloud inputs, achieves an accuracy of $59.2 \%$, $2.4 \%$ lower than the IMF-enhanced model, which achieves $61.6 \%$.
These results demonstrate that the IMF Module outperforms both the MLP-based and Attention-based fusion models across all datasets, highlighting its effectiveness in improving model accuracy and incomplete multimodal fusion.
}

We conduct a case study to compare existing large-scale Vision-Language Models (LVLMs) with City-VLM, which is shown in Figure~\ref{fig:case_study}.
Existing LVLMs, such as LLaVA and BLIP-2, often produce hallucination responses when addressing measurement questions. 
For example, when asked, "\textit{What is the situation of the barren land?}", these models may generate unreliable information. 
In contrast, our proposed model effectively reduces these hallucinations, providing more accurate and reliable answers to measurement queries.

\section{Conclusion}

In this work, we investigate the large vision language model (LVLM) for outdoor scene understanding from both dataset and method perspectives. 
Firstly, we introduce \textbf{SVM-City}, the first outdoor city-level multi\textbf{\underline{S}}cale, multi\textbf{\underline{V}}iew, and multi\textbf{\underline{M}}odal dataset to cover multidomain perception to profile cities.
Secondly, we propose \textbf{City-VLM}, an LVLM that constructs a joint probabilistic distribution space over 2D and 3D modalities, which enhances the visual representation for LVLM in case of corrupted sensor modalities. 
Our experimental results demonstrate that City-VLM outperforms existing LVLMs $18.14 \%$ on three outdoor QA tasks, demonstrating its superior ability to understand across various scene scales.
To our knowledge, we are the first to explore LVLMs based on multimodal incomplete learning, as well as their application in outdoor scene understanding, which can promote the development of human-environment interaction within outdoor scenes.

\section{Acknowledgments}

This work is supported by the following
programs: a Hong Kong RIF grant under Grant No. R6021-20; Hong Kong CRF grants under Grant No. C2004-21G and C7004-22G; the Postdoctoral Fellowship Program of CPSF under Grant Number GZC20232292.

\clearpage

\end{document}